\PassOptionsToPackage{table}{xcolor}
\documentclass[]{fairmeta}

\usepackage{silence}
\usepackage[utf8]{inputenc} 
\usepackage[T1]{fontenc}    
\usepackage{url}            
\usepackage{booktabs}       
\usepackage{amsfonts}       
\usepackage{nicefrac}       
\usepackage{microtype}      
\usepackage{xcolor}
\usepackage{soul}
\definecolor{bluelink}{RGB}{0,113,188}
\definecolor{greenlink}{RGB}{0,188,113}
\definecolor{PineGreen}{rgb}{0.0, 0.47, 0.44}
\definecolor{Gray}{rgb}{0.5,0.5,0.5}
\definecolor{audio_desc}{rgb}{0.82,0.99,0.80}
\definecolor{shot_desc_1}{rgb}{0.98,0.87,0.87}
\definecolor{shot_desc_2}{rgb}{0.98,0.95,0.83}
\definecolor{shot_desc_3}{rgb}{0.92,0.86,0.98}
\usepackage{listings} 
\usepackage{wrapfig}
\usepackage[most]{tcolorbox}

\newtcolorbox{samplebox}[1]{
    breakable, 
    colback=blue!5!white,
    colframe=blue!50!black, 
    fonttitle=\bfseries,
    title=#1
}
\definecolor{citecolor}{HTML}{0071bc}
\hypersetup{
    colorlinks=true,%
    citecolor=citecolor,%
    filecolor=citecolor,%
    linkcolor=citecolor,%
    urlcolor=citecolor
}
\usepackage{tabularx}
\usepackage{amsmath}
\usepackage{multirow}
\usepackage{array}
\usepackage{caption}
\usepackage{enumitem}
\usepackage{tikz}
\usepackage{lipsum}
\usepackage{adjustbox}
\usepackage{siunitx}
\usepackage{graphicx}     
\usepackage{float}

\usepackage{amssymb}
\renewcommand{\paragraph}[1]{\vspace{1.25mm}\noindent\textbf{#1}}

\usepackage{colortbl}       
\definecolor{oursrow}{RGB}{220, 235, 255}

\usepackage{algorithm}      
\usepackage{algorithmic}    

\definecolor{claimred}{RGB}{196, 48, 48}      
\definecolor{claimblue}{RGB}{0, 102, 179}     

\definecolor{ao}{rgb}{0.0, 0.0, 1.0}
\definecolor{airforceblue}{rgb}{0.36, 0.54, 0.66}
\definecolor{ceruleanblue}{rgb}{0.16, 0.32, 0.75}
\definecolor{cerulean}{rgb}{0.0, 0.48, 0.65}
\definecolor{celestialblue}{rgb}{0.29, 0.59, 0.82}
\definecolor{azure(colorwheel)}{rgb}{0.0, 0.5, 1.0}
\definecolor{babyblue}{rgb}{0.54, 0.81, 0.94}
\definecolor{babyblueeyes}{rgb}{0.63, 0.79, 0.95}
\definecolor{ballblue}{rgb}{0.13, 0.67, 0.8}

\definecolor{asparagus}{rgb}{0.53, 0.66, 0.42}
\definecolor{ao(english)}{rgb}{0.0, 0.5, 0.0}
\definecolor{applegreen}{rgb}{0.55, 0.71, 0.0}
\definecolor{armygreen}{rgb}{0.29, 0.33, 0.13}
\definecolor{gray-asparagus}{rgb}{0.27, 0.35, 0.27}
\definecolor{green(ryb)}{rgb}{0.4, 0.69, 0.2}

\definecolor{amethyst}{rgb}{0.6, 0.4, 0.8}
\definecolor{antiquefuchsia}{rgb}{0.57, 0.36, 0.51}
\definecolor{blue-violet}{rgb}{0.54, 0.17, 0.89}
\definecolor{brightlavender}{rgb}{0.75, 0.58, 0.89}
\definecolor{brightube}{rgb}{0.82, 0.62, 0.91}
\definecolor{brilliantlavender}{rgb}{0.96, 0.73, 1.0}

\definecolor{amber}{rgb}{1.0, 0.75, 0.0}
\definecolor{amber(sae/ece)}{rgb}{1.0, 0.49, 0.0}
\definecolor{atomictangerine}{rgb}{1.0, 0.6, 0.4}
\definecolor{burntorange}{rgb}{0.8, 0.33, 0.0}
\definecolor{burntsienna}{rgb}{0.91, 0.45, 0.32}
\definecolor{cadmiumorange}{rgb}{0.93, 0.53, 0.18}
\definecolor{carrotorange}{rgb}{0.93, 0.57, 0.13}
\definecolor{chocolate(web)}{rgb}{0.82, 0.41, 0.12}
\definecolor{chromeyellow}{rgb}{1.0, 0.65, 0.0}
\definecolor{darkorange}{rgb}{1.0, 0.55, 0.0}
\definecolor{darktangerine}{rgb}{1.0, 0.66, 0.07}
\definecolor{deepcarrotorange}{rgb}{0.91, 0.41, 0.17}
\definecolor{deepsaffron}{rgb}{1.0, 0.6, 0.2}
\definecolor{fulvous}{rgb}{0.86, 0.52, 0.0}

\newlength\savewidth
\newcolumntype{x}[1]{>{\centering\arraybackslash}p{#1pt}}
\newcolumntype{y}[1]{>{\raggedright\arraybackslash}p{#1pt}}
\newcolumntype{z}[1]{>{\raggedleft\arraybackslash}p{#1pt}}

\usepackage{pgf}
\usepackage{rotating}
\usepackage[abs]{overpic}
\usepackage{makecell}
\usepackage{longtable}
\usepackage{csquotes}
\definecolor{eventcolor}{RGB}{70,130,180}   
\definecolor{shotcolor}{RGB}{218,165,32}    
\definecolor{entitycolor}{RGB}{220,20,60}   
\definecolor{globalcolor}{RGB}{150,210,120}   

\title{HarmoView: Harmonizing Multi-View Constraints for Identity-Consistent Video Generation}
\newcommand{\papertitle}{HarmoView: Harmonizing Multi-View Constraints for Identity-Consistent Video Generation}

\author[]{Cong Wang}
\author[]{Zhentao Yu}
\author[]{Hongmei Wang}
\author[]{Weicong Liang}
\author[]{Zixiang Zhou}
\author[]{\\Zilin Yang}
\author[]{Jiarong Ou}
\author[]{Rui Chen}
\author[]{Yezhou Liu}
\author[]{Shiyu An}
\author[]{Yue Lu}
\author[]{Yuan Zhou}
\author[]{Qinglin Lu}
\affiliation[]{Tencent Hunyuan}

\usepackage{fancyhdr}
\fancypagestyle{titlepagewithlogo}{
  \fancyhf{} 
  \rhead{\includegraphics[height=0.6cm]{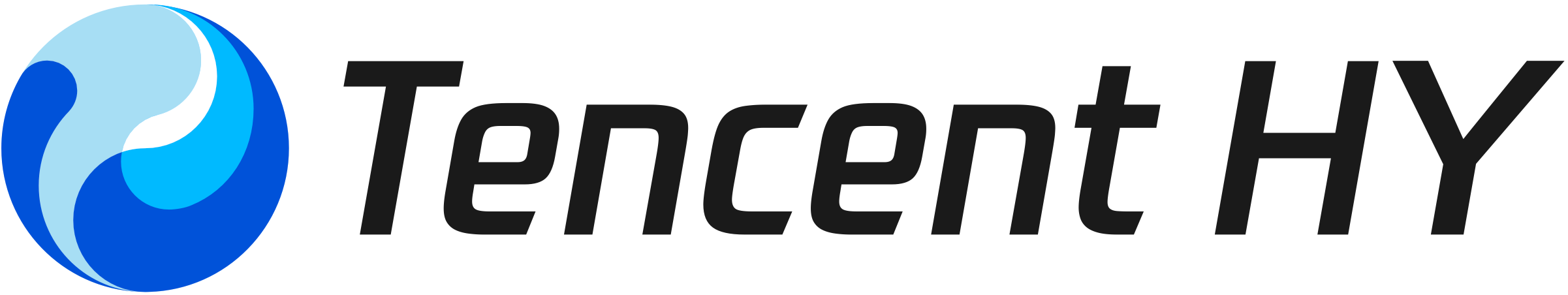}} 
}

\abstract{} 

\begin{document}

\renewcommand{\thefootnote}{\fnsymbol{footnote}}
\thispagestyle{titlepagewithlogo}
\maketitle
\footnotetext{Corresponding Email: cw\_research@163.com}
\renewcommand{\thefootnote}{\arabic{footnote}}
  

\begin{center}
{\large\textbf{Abstract}}
\end{center}
Current identity-consistent video generation methods struggle to preserve appearance fidelity under large viewpoint changes.
While introducing multi-view reference input offers a natural solution, progress remains constrained by the lack of effective frameworks for multi-view inputs and the scarcity of multi-view data.
We address these challenges by proposing \textbf{HarmoView}, a robust framework for identity-consistent video generation that effectively integrates multi-view cues through three synergistic architectural refinements complemented by a staged training curriculum.
Specifically, (1) we first introduce Multi-level Feature Injection (MFI) to anchor identity fidelity; by injecting raw ViT features from frontal reference images alongside text tokens via cross-attention, MFI provides persistent low-level appearance anchors that complement the high-level identity features within DiT blocks, leading to enhanced identity preservation.
Then, (2) we employ learnable proxy tokens to unify heterogeneous reference layouts across single- and multi-view settings while simultaneously resolving the reference--view mismatch problem.
(3) Jump-RoPE is further developed for identity-wise feature isolation to reduce identity crosstalk.
To activate these structural capabilities while preserving the model's original generative priors, we propose the Progressive View Curriculum. 
This four-stage training strategy employs view dropout to facilitate a stable transition from vanilla T2V generation to high-fidelity, identity-persistent spatial reasoning.
Furthermore, we construct a large-scale multi-view dataset to address the issue of data scarcity using an in-house LoRA-augmented pipeline and multi-stage filtering processes.
Extensive evaluation on our multi-view benchmark---comprising 100 manually-curated cases spanning 52 unique identities---demonstrates that HarmoView significantly outperforms open-source baselines and matches leading closed-source engines, achieving state-of-the-art performance in identity-consistent video generation.

\section{Introduction}

\begin{figure}[t]
    \centering
    \includegraphics[width=\linewidth]{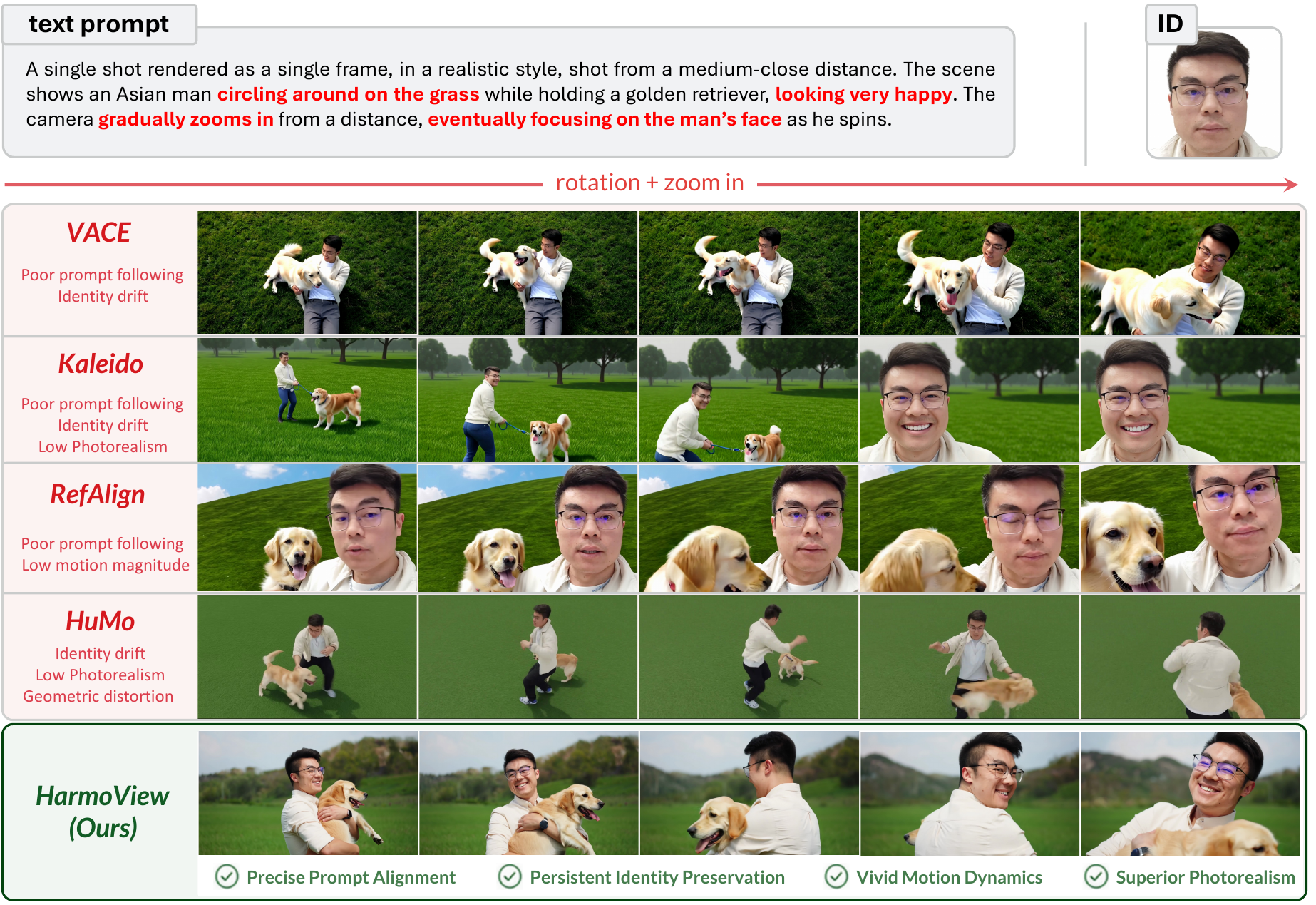}
    \caption{\textbf{Qualitative comparisons under large-viewpoint transitions.}
    In scenarios involving drastic rotations and zooming, existing methods typically suffer from identity drift and geometric distortion. 
    In contrast, HarmoView achieves superior identity persistence and structural integrity.
    }
    \label{fig:teaser}
\end{figure}

The pursuit of identity-consistent video generation—the ability to synthesize high-fidelity videos that strictly adhere to a target person's facial features—has witnessed significant breakthroughs with the emergence of Diffusion Transformers (DiTs) \citep{DBLP:journals/corr/abs-2502-11079, DBLP:journals/corr/abs-2503-07598, DBLP:journals/corr/abs-2505-04512, DBLP:journals/corr/abs-2509-08519, DBLP:journals/corr/abs-2603-25743}.
However, under large-angle synthesis, these models usually exhibit two characteristic failures: 
(1) hallucination, where unseen views are filled with fabricated facial structure unrelated to the target, and (2) reference–view mismatch, where frontal features are projected incorrectly onto a side view, thereby producing unnatural geometry or the ``mean face'' effect (an averaged, identity-less appearance, as shown in \cref{fig:teaser}). 
Both failures share a single root cause: \emph{a lack of view-specific appearance evidence the model can ground large-angle predictions in}.

A natural remedy is to condition on multiple reference views (e.g., a frontal plus two side views) so that large-angle predictions are anchored in view-specific evidence rather than extrapolated from a single frontal image.
Yet, effectively implementing this integration remains non-trivial due to two primary challenges. 
\emph{First}, there is no principled way to integrate multi-view cues into a pre-trained DiT backbone.
Naively concatenating multi-view references with video noise often triggers two failure modes: 
(1) Inter-identity attention crosstalk. When multiple identity subjects are injected, the model suffers from cross-view attention contamination, causing features from different identities to blend uncontrollably. 
(2) Training overhead induced by sequnce OOD (Out-Of-Distribution). The extended token sequences deviate from the pre-training distribution, forcing the model to require significantly larger datasets and longer convergence steps to maintain the video generation quality of the original backbone.
\emph{Second}, multi-view video–identity pairs are effectively absent in existing data. 
Web-scale video corpora contain only unconstrained, single-view footage; curated multi-view portrait datasets cover at most a few dozen subjects and lack the motion diversity required by a video generator.

Recent advancements in commercial video generation \citep{seedance2026, kling2026} have begun exploring multi-view identity injection for enhanced identity persistence. 
However, academic research in this domain remains remarkably sparse.
A concurrent work, MV-S2V \citep{DBLP:journals/corr/abs-2601-17756}, attempt to address multi-view subject-consistent generation but focuses primarily on rigid objects, proposing a data-synthesis pipeline and a set of RoPE-shift variants for cross-view positional encoding.
However, its assumptions break down on human portraits. Rigid objects admit sparse landmark correspondences and tolerate coarse cross-view alignment, whereas faces require sub-pixel precision in the eye–nose–mouth region to preserve identity and are more prone to cross-view feature blending.
Consequently, \emph{there is an urgent need for a public dedicated framework capable of effectively leveraging multi-view cues to ensure robust identity persistence.}

In this paper, we present \textbf{HarmoView}, a framework for identity-consistent video generation that effectively integrates multi-view reference cues through three synergistic architectural refinements complemented by a staged training curriculum, rectifying identity drift in large-angle synthesis.
Specifically, our architectural refinements address the bottleneck of identity persistence through three key innovations.
First, we introduce Multi-level Feature Injection (MFI) to anchor identity fidelity.
Unlike previous methods that rely solely on abstract identity latent from DiT blocks, MFI extracts additional visual features from frontal references and concatenates them with text tokens. 
By introducing this hybrid sequence with native cross-attention mechanism, the model can perceive these low-level appearance features throughout the entire denoising process, complementing the abstract high-level features for enhanced identity fidelity.
Second, learnable proxy tokens are employed as structural placeholders to unify heterogeneous reference layouts across single- and multi-view settings, which also act as ``attention sinks'' to absorb redundant features and mitigate reference--view mismatch problem within the single-view setting.
Third, we develop Jump-RoPE for identity-wise feature isolation. 
By introducing logical gaps between positional embedding bands of different identities, Jump-RoPE enforces strict feature boundaries, thus effectively suppress identity crosstalk.

Complementing these structural enhancements, the Progressive View Curriculum (PVC) ensures a stable evolution toward identity-persistent spatial reasoning. 
The curriculum begins with large-scale single-view continuous training to scale the base model (i.e., Wan2.2-T2V \citep{DBLP:journals/corr/abs-2503-20314}), followed by three joint training stages with strategic view dropout, forcing the model to internalize multi-view geometric relationships and preserving the generative quality of the original T2V priors. 

To support the model training, we also construct a large-scale multi-view dataset using a LoRA-augmented image-to-image synthesis pipeline followed by rigorous multi-stage filtering processes. 
Extensive evaluation on our HarmoView-Bench—comprising 100 manually-curated cases spanning 52 unique identities—demonstrates that HarmoView achieves state-of-the-art performance among open-source methods and comparable performance to leading closed-source methods in the domain of identity-consistent video generation.

The main contributions of this work are summarized as follows:
\begin{itemize}
    \item We propose \textbf{HarmoView}, the \emph{first} public human-centric framework to integrate multi-view reference cues into a pre-trained DiT video backbone with three synergistic architectural refinements for enhanced identity fidelity, especially under large viewpoint changes.

    \item We design the Progressive View Curriculum, a four-stage schedule with view dropout, that enables a stable transition to multi-view spatial reasoning without sacrificing the text-following and motion-synthesis capabilities of the underlying T2V backbone.

    \item We design a LoRA-augmented synthesis pipeline and rigorous multi-stage filtering processes, enabling the automated production of large-scale multi-view video-identity pairs.
    
    \item Extensive experiments demonstrate that HarmoView effectively leverages multi-view cues to significantly enhance identity-consistent video generation. On our curated benchmark, HarmoView achieves state-of-the-art performance among open-source methods and delivers results comparable to leading closed-source commercial engines.
\end{itemize}

\section{Related Work}

\noindent\textbf{Identity-consistent video generation.}
This field has evolved from audio-driven human animation, which focuses on lip-syncing and motion transfer relative to a fixed start frame, to more generalized identity-consistent video generation. 
While early animation methods \citep{DBLP:conf/cvpr/0003LZSCMMZ0Z25, DBLP:conf/mm/WangWJFZQZX25, DBLP:journals/corr/abs-2502-01061, DBLP:journals/corr/abs-2505-23525} achieve impressive realism, they remain fundamentally anchored to an identity-complete source frame, which simplifies appearance consistency but limits creative flexibility under large viewpoint shifts.
Modern approaches bridge this gap by treating reference images as identity priors within a Text-to-Video backbone. 
Existing strategies typically employ external adapters \citep{DBLP:journals/corr/abs-2404-15275} or in-context methods \citep{DBLP:journals/corr/abs-2502-11079, DBLP:journals/corr/abs-2503-07598, DBLP:journals/corr/abs-2505-04512, DBLP:journals/corr/abs-2509-08519} that concatenate reference latents with noisy video latents.
Although this leverages the inherent spatial consistency of pre-trained DiTs, it often risks weakening textual control or suffering from identity drift under large-viewpoint generation.
Unlike prior methods that rely on a single frontal image and struggle to preserve fine-grained identity details, HarmoView explicitly incorporate multi-view cues within a unified framework, achieving enhanced identity persistence and video naturalness.

\noindent\textbf{Multi-view conditioning.}
While multi-view injection has been extensively explored in fields such as 3D reconstruction and novel-view synthesis to enhance geometric performance \citep{DBLP:conf/cvpr/CaoYL0X025, DBLP:conf/cvpr/WangKSQWBZ25, DBLP:journals/corr/abs-2509-06579, sun2025bootstrap3d}, its application in identity-consistent video generation remains underdeveloped. 
In the current landscape, multi-view capabilities are primarily restricted to closed-source commercial engines \citep{seedance2026, kling2026, wan2026, vidu2026} or specialized frameworks like MV-S2V \citep{DBLP:journals/corr/abs-2601-17756}, which predominantly focus on the generation of rigid objects with well-defined geometries.
HarmoView represents the first public effort to explicitly introduce multi-view cues into the realm of identity-consistent human video generation

\begin{figure}[t]
    \centering
    \includegraphics[width=\linewidth]{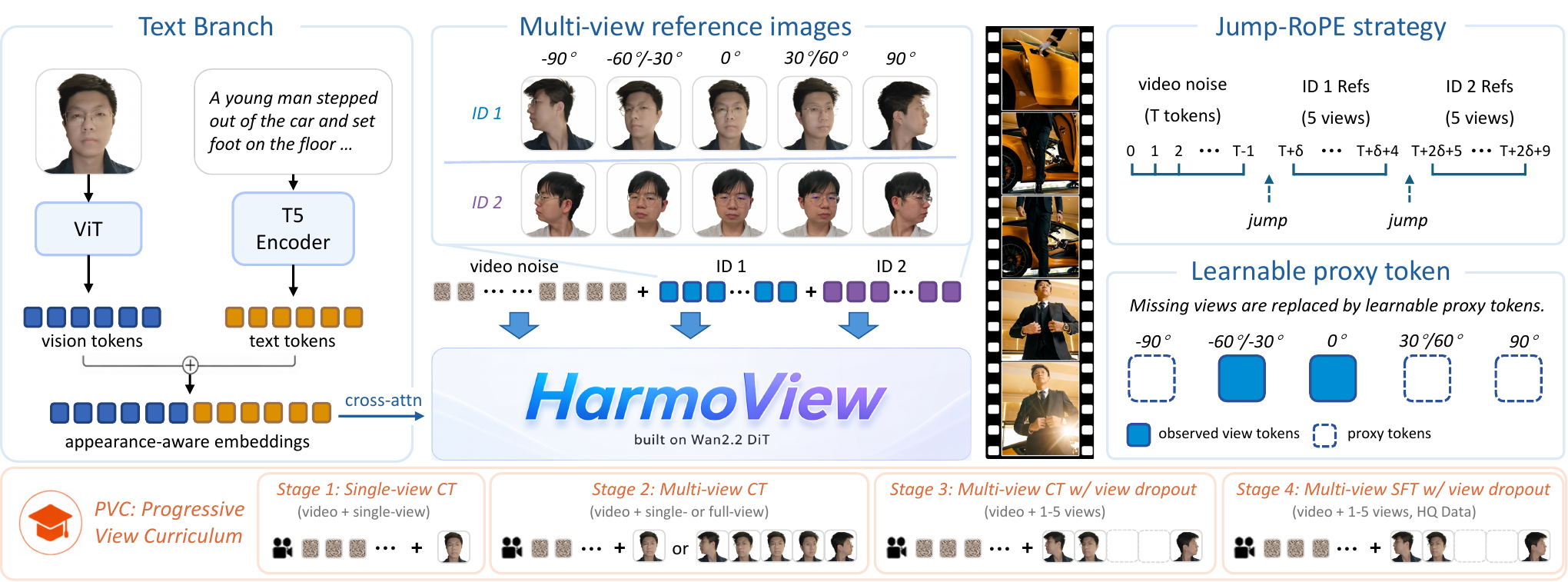}
    \caption{\textbf{Overview of HarmoView.} Our framework effectively integrates multi-view cues to pre-trained Wan2.2-T2V model through three key architectural refinements for enhanced identity persistence, complemented by a four-stage progressive view curriculum for stable training.
    }
    \label{fig:pipeline}
\end{figure}

\section{The HarmoView Framework}
\label{sec:harmoview}
Built upon the Wan2.2-T2V architecture, our HarmoView framework (\cref{fig:pipeline}) synthesizes identity-consistent videos by integrating two primary information streams: descriptive text tokens injected via cross-attention, and a set of multi-view reference images $\{I_{ref}^i\}_{i=1}^N$ that are encoded into latents and concatenated with the video noise along the sequence dimension . 
To effectively leverage multi-view cues without compromising pre-trained generative priors, our method introduces three targeted architectural refinements (\cref{sec:arch_refine}), complemented by the Progressive View Curriculum (PVC, \cref{sec:pvc}) to ensure a stable transition to multi-view constraints.
Specifically, we first propose Multi-level Feature Injection (MFI) to anchor global identities by embedding fine-grained appearance features directly into the text conditioning sequence, ensuring these cues remain perceptible throughout the denoising process.
Then, learnable proxy tokens are introduced to unify the sequence layout across heterogeneous references while functioning as "attention sinks" to mitigate reference--view mismatch problem.
To further suppress feature crosstalk between distinct identities, we implement Jump-RoPE that introduces logical gaps in the rotary positional embeddings to enforce separation between different identity tokens.
These structural capabilities are systematically activated via the PVC, a four-stage training strategy that facilitates the learning of multi-view spatial reasoning without degrading the model's original generative quality.
Finally, to provide a large-scale, high-fidelity foundation for effective training and benchmarking, we develop an automatic data pipeline (\cref{sec:datapipe}), including a LoRA-augmented synthesis engine and multi-stage filtering processes.

\subsection{Architectural Refinements for Identity Persistence}
\label{sec:arch_refine}
\subsubsection{Multi-level Feature Injector}
\label{sec:mfi}
While DiTs excel at modeling high-level spatial relationships, their layer-wise abstraction often leads to the gradual erosion of fine-grained identity details. 
To counteract this loss and provide the model with persistent detailed guidance, we propose MFI.
Instead of relying on complex latent space mapping between images and text, MFI enriches the conditioning sequence by surfacing low-level appearance features directly within the text branch. 
Specifically, given a canonical frontal reference, we extract visual tokens $v_{img} \in \mathbb{R}^{L_v \times D}$ using a frozen ViT encoder \citep{DBLP:journals/corr/abs-2502-13923}. 
These tokens are then projected via a single linear layer and concatenated with textual tokens along the sequence dimension, forming a unified conditioning sequence:
{
\setlength{\abovedisplayskip}{3pt}
\setlength{\belowdisplayskip}{3pt}
\begin{equation}
z_{mm} = [\text{Linear}(v_{img}); \text{T5}(y)]
\end{equation}
}
The core insight behind MFI is to leverage multi-level feature supervision to enhance identity consistency.
While the DiT blocks transform reference cues into high-level spatial representations to capture geometric structure, MFI preserves raw low-level appearance priors within the conditioning buffer.
By injecting these low-level features as constant, high-fidelity anchors that bypass the abstraction layers of the DiT, we ensure that fine-grained facial details are perceptible throughout the entire denoising process.
Instead of employing auxiliary cross-attention heads, which often introduce feature competition and additional computation cost, our concatenation strategy leverages the model's existing attention mechanism to perform joint spatial-semantic reasoning, forcing the model to resolve their inter-dependencies within a consistent and mutually-aware attention map.

Furthermore, we utilize only frontal references as the anchor of identities, which strikes an optimal balance between providing a sufficient guidance and preventing the attention weight dilution typically caused by extended sequences, thereby preserving the model's strict adherence to textual instructions.

\subsubsection{Learnable Proxy Tokens}
Within the DiT branch, we encode the reference images $\{I_{ref}^i\}_{i=1}^N$ into latents and concatenate them with the video noise along the sequence dimension. 
However, since the number of available reference views can vary between training stages and inference scenarios, directly feeding these variable-length tokens into the DiT blocks would disrupt the structural priors of the computation graph.

To address this, we introduce a learnable proxy token $\mathcal{T}_{proxy} \in \mathbb{R}^{P^2 \times C}$, where $P \times P$ represents the spatial resolution of the latent patches. 
These tokens are randomly initialized and optimized alongside the model parameters.
When encountering a missing view (e.g., during single-view training or scenarios with limited reference views), we replicate $\mathcal{T}_{proxy}$ to match the latent resolution of a standard reference image and place these expanded tokens at the positions of the absent views.
This design serves a dual role:

\noindent\textbf{Layout Unification and Semantic Consistency.}
By fixing the number of reference slots (e.g., $N=5$), we ensure a unified token sequence layout across all stages. 
This provides deterministic positional semantics, allowing the DiT backbone to internalize robust spatial priors without the instability of dynamic sequence lengths, benefiting both joint training and inference phases.

\noindent\textbf{Attention Sinks for Mismatch Mitigation.}
The proxy tokens further resolve the reference–view mismatch problem.
For example, when synthesizing a side-view video from only a frontal reference, due to the utilization of the softmax activation, the attention mechanism usually forces a false matching with irrelevant features due to the lack of correct spatial correspondence.
Functioning as attention sinks, these proxy tokens learn to absorb these irrelevant weights, providing a dedicated space for unmappable features.
This prevents conflicting identity cues from contaminating the video latents, ensuring that HarmoView not only effectively leverages multi-view cues when available but also strengthens the single-view reference capability.

\subsubsection{Jump-RoPE}
\label{sec:jump_rope}
Despite structural unification, the model suffers from identity crosstalk due to the inherent continuity of Rotary Positional Embeddings (RoPE). 
This positional leakage often leads to spatially inconsistent identity blending, where the model erroneously assembles disjointed facial features from different identities, such as blending the upper-face of one identity with the jawline structure of another.

To resolve this, inspired by MV-S2V, we introduce \textbf{Jump-RoPE}, which implements discontinuous coordinate jumps to achieve clean feature decoupling. 
Unlike standard sequential indexing, Jump-RoPE applies two distinct levels of positional offsets:
\textbf{(1) Inter-identity Isolation.} 
For scenarios involving multiple reference identities, we insert a significant index gap between the token blocks of different individuals, making their respective features occupy distant regions in the rotation space and alleviating cross-identity attribute leakage.
\textbf{(2) Reference-to-Video Decoupling.} 
Another index jump is further applied between the entire reference sequence and the video noise tokens to identify the role of different RoPE bands and achieve logical isolation.

Jump-RoPE ensures that the reference and video latents are situated in logically distant regions of the high-dimensional coordinate system, effectively circumvents the identity crosstalk problem.

\subsection{Progressive View Curriculum}
\label{sec:pvc}
The design of the Progressive View Curriculum (PVC) is necessitated due to the extreme scarcity of high-quality multi-view identity-video pairs.
Unlike a ``brute-force'' approach that risks over-fitting or collapsing the model's pre-trained priors due to data limitations, PVC utilizes a four-stage optimization schedule that gradually increase the complexity of reference cues.
This design ensures a stable transition toward multi-view spatial reasoning, while preserving the text-following and motion-synthesis capabilities of the underlying T2V backbone.

\noindent\textbf{Stage 1: Single-View Structure Alignment and Capacity Scaling.}
This stage aims to align the pre-trained T2V backbone with our architectural refinements by leveraging the abundance of single-view video-identity pairs. 
Through continuous training, the model learns to anchor identity semantics and establish a high-fidelity baseline for identity persistence using single-view references. 
This foundational step ensures the structural modifications are fully integrated and the model's  capacity on identity persistence is scaled up before introducing the geometric complexities of multi-view fusion.

\noindent\textbf{Stage 2: Multi-View Alignment and Knowledge Injection.} 
Building upon the single-view consistency from Stage 1, this stage introduces multi-view cues to establish a fundamental mapping between multi-view inputs and target video latents.
We employ a dual-configuration training strategy, utilizing either single-view or complete full-view (all $N$ references) inputs.
By bypassing the complexity of missing-view scenarios, this "warm-up" phase reduces the learning difficulty of multi-view integration, enabling the model to effectively capture global spatial correspondences.

\noindent\textbf{Stage 3: Strategic View Dropout for Spatial Reasoning.}
To enhance the model's capacity for sparse conditioning, this stage introduces a strategic view dropout mechanism where we randomly sample a variable number of reference views (from 1 to $N$). 
Rather than just improving robustness, this strategy compels the model to internalize precise positional semantics by forcing it to rely on specific geometric cues from dynamic active slots, ensuring global coherence and accurate spatial reasoning even under imperfect or limited conditioning.

\noindent\textbf{Stage 4: High-Fidelity Refinement and Identity Polishing.}
The final stage serves as a polishing phase, shifting the optimization focus from structural alignment to aesthetic refinement and fine-grained identity reconstruction.
By utilizing a curated, high-quality data subset while reducing the intensity of view dropout, we enable the model to prioritize the reconstruction of subtle facial textures and idiosyncratic features.
This phase effectively bridges the gap between raw multi-view consistency and professional-grade visual quality, seamlessly integrating the previously learned spatial reasoning with the pixel-level fidelity required for state-of-the-art identity-consistent video generation.

\begin{figure}[t]
    \centering
    \includegraphics[width=\linewidth]{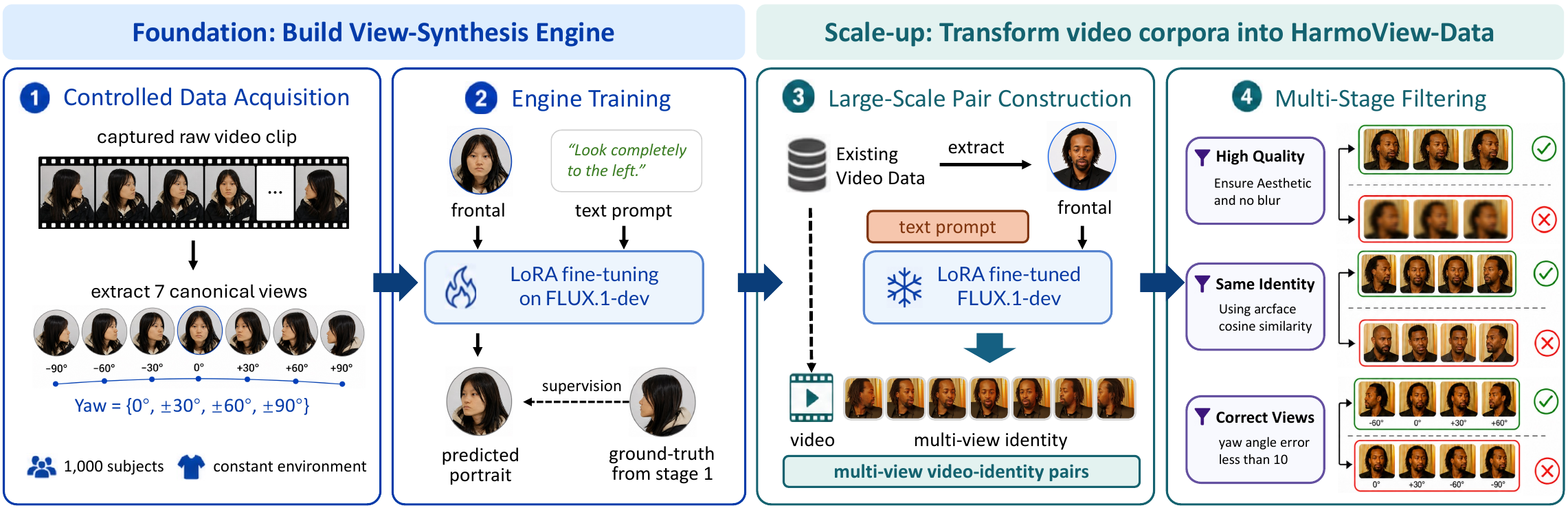}
    \caption{\textbf{Scalable Multi-View Data Pipeline.} Our automated data pipeline leverages a LoRA-augmented synthesis engine and a multi-stage filtering protocol to construct high-quality multi-view identity-video pairs for effective training and benchmarking.}
    \label{fig:data_pipeline}
\end{figure}

\subsection{Scalable Multi-View Data Pipeline}
\label{sec:datapipe}
To overcome the extreme scarcity of high-quality multi-view video-identity pairs, we develop an automated pipeline to construct the HarmoView-Dataset.
The pipeline leverages an in-house LoRA-augmented view-synthesis engine that supports transforming massive existing video corpora into multi-view video-identity pairs, complemented by a multi-stage filtering protocol to ensure geometric accuracy and identity persistence.

Specifically, to establish a foundational mapping for viewpoint manipulation, we distill precise multi-view geometric priors from a curated high-fidelity dataset into a generative framework.

\noindent\textbf{Controlled Data Acquisition.} 
We first curated a proprietary dataset of 1,000 diverse subjects performing standardized head-rotation sequences. 
Using the InsightFace framework \citep{insightface}, we extract seven canonical frames per subject at specific yaw angles $Y = \{0^\circ, \pm 30^\circ, \pm 60^\circ, \pm 90^\circ\}$. 
Since frames are sampled from a single continuous clip, all environmental factors (e.g., lighting and background) remain strictly constant, effectively isolating head orientation as the sole variable.

\noindent\textbf{Engine Training.} 
Using these aligned sets, we fine-tune a LoRA adapter on the FLUX.1 model \citep{DBLP:journals/corr/abs-2506-15742} to develop a specialized view-synthesis engine. 
By framing the task as a conditional generation problem where the input is a frontal reference and the target is a portrait in specific yaw angles, the engine learns to synthesize high-fidelity portraits at specific orientations, acting as a "pseudo-labeler," allowing us to project any frontal portrait image into multi-view sets while perseving identity features.

With the view-synthesis engine established, we construct our multi-view training data by transforming existing raw videos into multi-view video-identity pairs.
For each video in the corpus, we first extract a high-quality frontal reference frame $I_{ref}$ using existing methods. 
This reference is then processed by the synthesis engine to generate a set of multi-view portraits $\{I_{i}\}_{i=1}^N$ based on given yaw angles, which are paired with the original video to form the required training pairs.

To ensure the generated data meets the high standards of our training framework, we implement a multi-stage filtering protocol to verify the quality of these synthetic pairs through a sequence of rigorous checks:
First, we apply (1) image quality filtering by utilizing an automated aesthetic predictor \citep{DBLP:journals/tip/TalebiM18} and a blur-detection operator to filter out low-quality generations, ensuring only ``pixel-perfect'' images proceed to subsequent verification.
This is followed by (2) identity verification, where we employ the InsightFace to calculate the cosine similarity between the synthesized views and the original reference frame. 
To account for the inherent difficulty of cross-view identity matching, we apply a viewpoint-dependent thresholding strategy, gradually relaxing the similarity requirement as the yaw angle increases from the frontal pose.
Finally, we perform (3) geometric validation using a pose estimator \citep{headposeestimation} to verify the head poses of the synthesized images, ensuring the target yaw angles are accurately manifested within a $\pm 10^\circ$ tolerance.
Through this rigorous orchestration, we successfully bridge the gap between massive video data and the large-scale requirements of our multi-view spatial reasoning framework.

\begin{figure}[t]
    \centering
    \includegraphics[width=\linewidth]{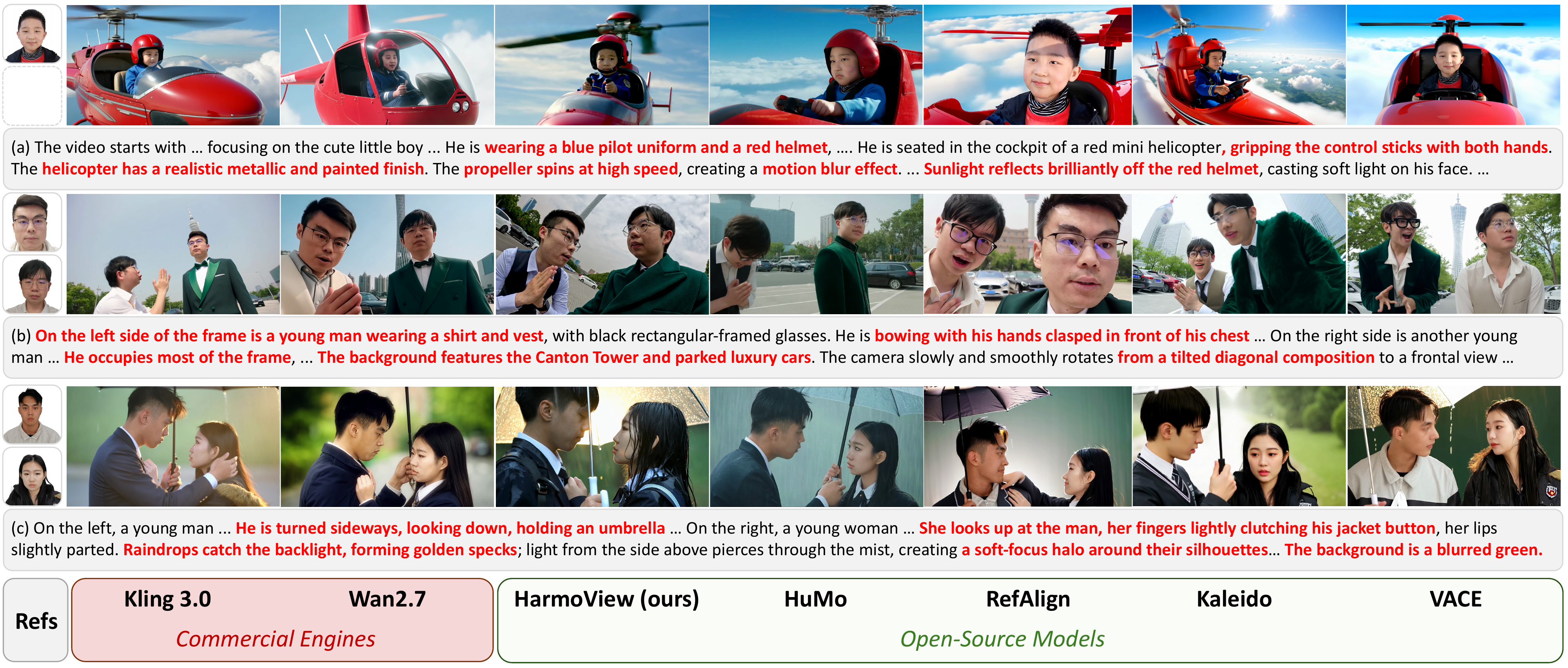}
    \caption{\textbf{Comparison with State of the Art Methods.}}
    \label{fig:main_comp}
\end{figure}

\section{Experiments}
\subsection{Experimental Setup}
\noindent\textbf{HarmoView-Bench.}
Due to the lack of existing video generation benchmarks that support multi-view conditioning, we introduce \textbf{HarmoView-Bench}, a manually curated evaluation suite comprising 100 diverse human-centric scenarios.
Each test case involves up to two distinct identities, with each identity accompanied by seven canonical yaw-angle reference images: $\{0^\circ, \pm 30^\circ, \pm 60^\circ, \pm 90^\circ\}$.
These scenarios encompass a broad spectrum of skin tones and complex motion trajectories, specifically designed to test a model's ability to aggregate disjointed spatial cues across multiple perspectives.
Detailed benchmark descriptions and visual examples are provided in the Supplementary Materials.
%
%

\noindent\textbf{Evaluation Metrics.}
To ensure an objective and authoritative assessment, we adopt the comprehensive evaluation framework from OpenS2V-Eval \citep{DBLP:journals/corr/abs-2505-20292}, a standardized benchmark for human-domain video generation. 
We evaluate the generated videos across three primary axes:
\begin{itemize}
    \item \textbf{Video Quality}: We assess the aesthetic appeal and temporal coherence of the videos. 
    This includes measuring visual beauty via aesthetic models (AES), temporal stability via motion smoothness metrics (Motion Smooth), physical realism through naturalness scoring (Natural Score), and motion amplitude (Motion Amp).
    \item \textbf{Text Alignment}: To evaluate adherence to input prompts, we utilize multimodal similarity scores (GME) based on MLLM, ensuring the generated motion and environment accurately reflect the textual descriptions.
    \item \textbf{Identity Consistency}: 
    We employ ArcFace-based facial similarity (FaceSim) to verify human identity persistence.
    Additionally, the Nexus score, which calculates the similarity between references and detected subject crops, is employed to mitigate background noise.
\end{itemize}
To complement these objective metrics and account for potential biases in automated scoring, we further conduct a \textbf{GSB (Good/Same/Bad)} human preference study. 
For each test case, three independent annotators perform side-by-side comparisons between HarmoView and competing models, with the final rating determined by a majority vote. 
By involving this subjective preference study, we provide a balanced and rigorous assessment of the model’s performance.

\noindent\textbf{Baselines.}
We compare HarmoView against a broad range of state-of-the-art methods to demonstrate its superiority in both open-source and commercial landscapes. 
The baselines include: (i) Open-source academic models, including VACE \citep{DBLP:journals/corr/abs-2503-07598}, Kaleido \citep{DBLP:journals/corr/abs-2510-18573}, RefAlign \citep{DBLP:journals/corr/abs-2603-25743}, and HuMo \citep{DBLP:journals/corr/abs-2509-08519}; and (ii) Leading commercial engines, including Wan 2.7 \citep{wan2026} and Kling 3.0 \citep{kling2026}. 
This selection ensures a comprehensive evaluation against the current frontier of human-centric video generation.

\begin{table}[t]
    \centering
    \vspace{-1.2em}
    \caption{Quantitative comparison on the HarmoView-Bench. All metrics are calculated following the OpenS2V-Eval protocol. \textbf{$^\star$} utilizes only the frontal reference.}
    \label{tab:main_results}
    \resizebox{0.95\linewidth}{!}{
    \begin{tabular}{l|c|cccc|c|cc}
    \toprule
    \multirow{2}{*}{\textbf{Method}} & \multirow{2}{*}{\textbf{Total $\uparrow$}} & \multicolumn{4}{c|}{\textbf{Video Quality}} & \textbf{Text Align.} & \multicolumn{2}{c}{\textbf{Identity Consistency}} \\
    &  & \textbf{Natural $\uparrow$} & \textbf{AES $\uparrow$} & \textbf{Smooth $\uparrow$} & \textbf{Motion Amp $\uparrow$} & \textbf{GME $\uparrow$} & \textbf{FaceSim $\uparrow$} & \textbf{Nexus $\uparrow$} \\
    \midrule
    \rowcolor[HTML]{E6F3E6} \multicolumn{9}{l}{\textit{Open-source Academic Models}} \\
    VACE & 67.74 & 73.64 & 54.10 & 91.21 & 26.70 & 60.01 & 42.84 & \textbf{98.17} \\
    Kaleido & 61.11 & 60.37 & \textbf{57.35} & \textbf{96.02} & 23.81 & 59.84 & 25.32 & 94.83 \\
    RefAlign & 68.71 & 65.14 & 50.10 & 92.44 & 30.35 & 60.90 & \textbf{60.13} & 97.51 \\
    HuMo & 73.11 & 83.67 & 53.15 & \underline{95.81} & 22.88 & \textbf{62.32} & \underline{59.69} & 94.52 \\
    \textbf{HarmoView$^\star$} & \underline{73.54} & \textbf{89.46} & \underline{54.89} & 89.80 & \textbf{38.57} & \underline{62.28} & 54.40 & 93.90 \\
    \textbf{HarmoView (Ours)} & \textbf{74.70} & \underline{89.12} & 53.80 & 91.68 & \underline{33.63} & 62.07 & 57.91 & \underline{97.89} \\
    \midrule
    \rowcolor[HTML]{FFF3E6} \multicolumn{9}{l}{\textit{Commercial Closed-source Engines}} \\
    Wan 2.7 & 73.53 & 84.86 & 59.44 & 97.69 & 16.28 & 61.43 & 61.12 & 89.35 \\
    Kling 3.0 & 70.92 & 78.57 & 50.86 & 89.02 & 23.75 & 59.64 & 57.88 & 96.87 \\
    \bottomrule
    \end{tabular}
    }
\end{table}

\subsection{Comparison with State-of-the-Art Methods}
We conduct comprehensive comparisons on the HarmoView-Bench against both academic baselines and industry-leading engines. 
As summarized in \cref{tab:main_results}, HarmoView achieves the best total score of 74.70, outperforming all open-source academic baselines and competitive commercial engines.
Notably, HarmoView exhibits a significant advantage in Video Quality, particularly in the Natural metric ($89.12$), which evaluates the overall realism and physical plausibility of the generated content. 
Compared to academic models like HuMo ($83.67$) and RefAlign ($65.14$), our method produces videos with fewer geometric artifacts under large-viewpoint transitions (see \cref{fig:main_comp}).
%

Furthermore, we observe that existing automated evaluation pipelines may introduce systematic biases in two key aspects.
First, identity metrics like FaceSim that relies on arcface-based embeddings exhibit a clear preference for frontal views where facial features are maximally visible.
As HarmoView generates a wider range of profile perspectives during large rotations, its objective identity score is inherently penalized compared to models that favor static, frontal outputs.
Second, there is often an inverse correlation between Motion Amp and overall aesthetic or smoothness scores in automated suites, where high-magnitude motion is frequently misinterpreted as a loss of stability.

To address these potential inaccuracies and provide a more authoritative assessment, we further conduct a human preference study (GSB) comparing HarmoView with six representative methods.
The results in \cref{fig:gsb_jump}(a) demonstrate that HarmoView consistently outperforms all six representative methods, including both open-source baselines and leading commercial engines, in human preference. 
This clear margin across all comparisons validates the superior efficacy of our approach in maintaining robust identity consistency and high generation quality.

\begin{table}[t]
    \centering
    \caption{Ablation analysis of HarmoView components on HarmoView-Bench. All models are evaluated for identity similarity and temporal stability.}
    \label{tab:ablation}
    \resizebox{0.95\linewidth}{!}{
    \begin{tabular}{c|l|c|cccc|c|cc}
    \toprule
    \multirow{2}{*}{\textbf{Label}} & \multirow{2}{*}{\textbf{Method}} & \multirow{2}{*}{\textbf{Total $\uparrow$}} & \multicolumn{4}{c|}{\textbf{Video Quality}} & \textbf{Text Align.} & \multicolumn{2}{c}{\textbf{Identity Consistency}} \\
     &  &  & \textbf{Natural $\uparrow$} & \textbf{AES $\uparrow$} & \textbf{Smooth $\uparrow$} & \textbf{Motion Amp $\uparrow$} & \textbf{GME $\uparrow$} & \textbf{FaceSim $\uparrow$} & \textbf{Nexus $\uparrow$} \\
    \midrule
    \rowcolor[HTML]{F2F2F2} -- & \textbf{HarmoView (Ours)} & \textbf{74.70} & \underline{89.12} & \textbf{53.80} & 91.68 & {33.63} & \textbf{62.07} & \textbf{57.91} & \textbf{97.89} \\
    \midrule
    (a)   & w/o MFI & 73.62 & \textbf{89.46} & 53.68 & 91.48 & 32.21 & 61.59 & 53.10 & 97.10 \\
    \midrule
    (b.1) & w/o LPT (Sequential Concat) & 70.41 & 80.50 & 50.45 & \textbf{92.80} & {34.25} & 61.40 & 51.50 & 95.50 \\
    (b.2) & w/o LPT (Gray Padding) & 72.46 & 82.81 & 51.76 & \underline{92.26} & \underline{35.11} & {61.46} & 56.31 & 97.15 \\
    \midrule
    (c) & w/o Jump-RoPE & \underline{74.51} & 89.10 & \underline{53.75} & 91.65 & 33.50 & \underline{62.05} & 56.80 & 97.75 \\
    \midrule
    (d.1) & only single-view training & 72.61 & 85.35 & 52.10 & 88.50 & \textbf{39.20} & 60.15 & 55.86 & 96.53 \\
    (d.2) & only multi-view training & 68.26 & 75.65 & 49.30 & 86.20 & 28.50 & 58.80 & 52.20 & 94.90 \\
    (d.3) & PVC w/o view dropout & \underline{74.51} & 88.85 & \underline{53.75} & 91.60 & 32.10 & 61.95 & \underline{57.25} & \underline{97.80} \\
    \bottomrule
    \end{tabular}
    }
\end{table}
    
\subsection{Ablation Studies}
To investigate the effects of each proposed component, we conduct a series of ablation experiments. 
Extended qualitative results are provided in our Supplementary Materials.

\begin{figure}[t]
    \centering
    \includegraphics[width=0.95\linewidth]{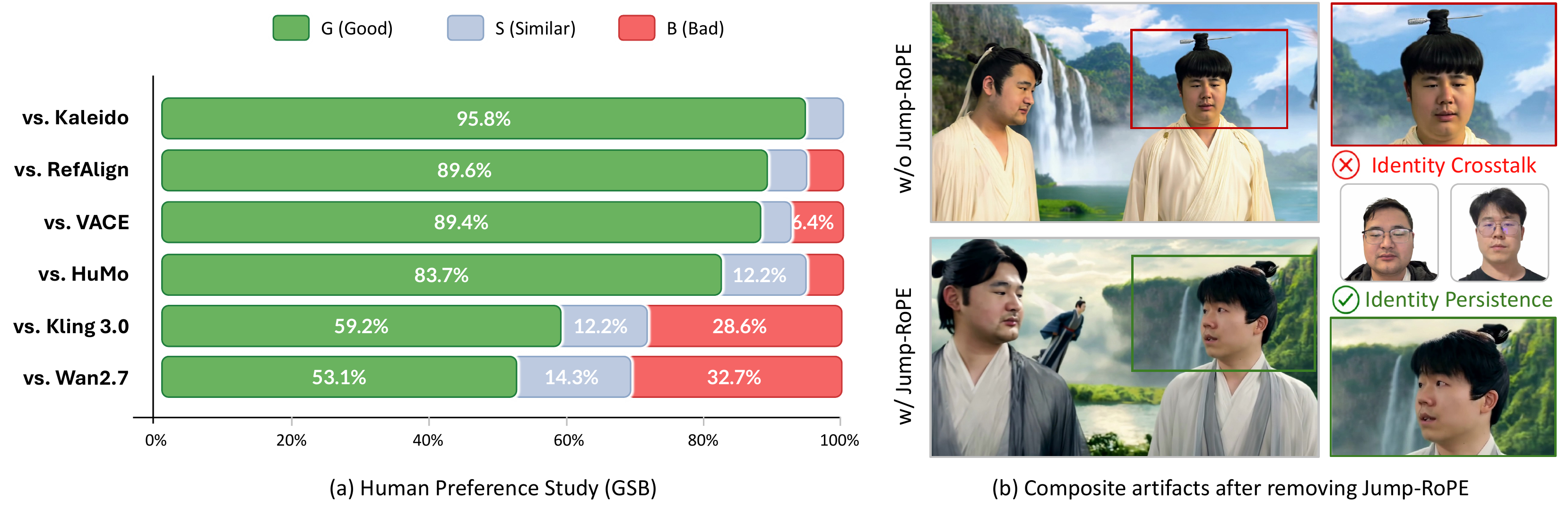}
    \caption{\textbf{Human Preference Study \& Ablation Study on Jump-RoPE.}}
    \label{fig:gsb_jump}
\end{figure}

\noindent\textbf{Efficacy of MFI.}
\cref{tab:ablation}(a) demonstrate that removing the MFI leads to a noticeable drop in the Total score from $74.70$ to $73.62$.
The most significant impact is observed in FaceSim (53.1 vs. 57.91), indicating that anchoring low-level identity features throughout the entire generation process is crucial for the enhanced identity persistence.
%
%

\noindent\textbf{Functionality of Learnable Proxy Tokens.}
We further investigate the design of Learnable Proxy Tokens (LPT) by comparing with two alternative padding strategies for missing views (\cref{tab:ablation}(b.1) and (b.2)).
Specifically, simply concatenating available views (Sequential Concat) results in a significant performance drop, with the Total score falling to $70.41$, as the lack of spatial anchors disrupts the fixed correspondence between RoPE-based position embeddings and specific viewpoints.
Replacing missing views with Gray Padding partially recovers performance ($72.46$ Total), yet it still lags behind our full model, particularly in terms of Naturalness ($82.81$ vs. $89.12$).
%
%

\noindent\textbf{Necessity of Jump-RoPE.}
\cref{tab:ablation}(c) shows that removing Jump-RoPE results in a marginal decrease in the Total score ($74.51$). 
However, without the positional gaps introduced by Jump-RoPE, the latent representations of different identities tend to suffer from severe crosstalk. 
This leads to composite artifacts where the generated subject inherits mismatched features from multiple references (e.g., \cref{fig:gsb_jump}(b) shows a blending between the jawline of Identity A and the upper-face of Identity B.)

\noindent\textbf{Impact of Progressive View Curriculum (PVC).}
Finally, we analyze the PVC by comparing it against three naive training baselines (\cref{tab:ablation}(d.1)--(d.3)).
When trained exclusively on single-view data (only single-view training), the model retains the strongest motion priors, achieving the highest Motion Amp score ($39.20$).
However, this comes at the cost of geometric consistency, as evidenced by the lower FaceSim ($55.86$) and Nexus ($96.53$) scores compared to the full model.
Conversely, training only on multi-view data (only multi-view training) results in poor convergence and the lowest Total score ($68.26$), primarily due to the limited scale of multi-view datasets relative to general video corpora.
While the variant without view dropout (PVC w/o view dropout) yields competitive results, it remains slightly inferior to our full framework in identity preservation.

\section{Conclusion}
In this work, we presented HarmoView, a novel framework designed to achieve high-fidelity, identity-consistent video generation through multi-view conditioning. 
\textbf{Our contribution is threefold}, spanning architecture, training strategy, and data engineering.
First, we introduced three architectural refinements: the Multi-level Feature Injection for global identity anchoring, Learnable Proxy Tokens for layout unification and mismatch mitigation, as well as the Jump-RoPE for physical feature isolation.
Second, recognizing the scarcity of multi-view training pairs, we proposed the Progressive View Curriculum that leverages limited multi-view data for stable transition from the T2V backbone to complex multi-view reasoning.
Finally, we developed a scalable multi-view data pipeline that bridges the gap between massive video data and the requirements of multi-view video-identity pairs. 
Extensive evaluations on our HarmoView-Bench demonstrate that HarmoView significantly outperforms existing academic baselines and remains highly competitive with top-tier commercial engines  in the domain of identity-consistent video generation.
%
%
We hope our framework and benchmark will catalyze further research into controllable and identity-persistent digital human generation.


\clearpage
\newpage

\label{references}
\addcontentsline{toc}{section}{References}
\bibliographystyle{assets/plainnat}
\bibliography{paper}

\end{document}